\documentclass[11pt,a4paper]{article}
\usepackage[hyperref]{acl2020}
\usepackage{times}
\usepackage{latexsym}
\usepackage{graphicx} 
\usepackage{url}
\usepackage{multirow}
\usepackage{color}
\usepackage{amsmath}
\usepackage{amssymb}

\usepackage{microtype}

\aclfinalcopy 
\def\aclpaperid{892} 

\title{Modeling Long Context for Task-Oriented Dialogue State Generation}

\author{Jun Quan  \and  Deyi Xiong\thanks{\ \  Corresponding author}\\
  \ \  School of Computer Science and Technology, Soochow University, Suzhou, China \\
  {\tt terryqj0107@gmail.com},\quad {\tt dyxiong@suda.edu.cn}}

\date{}

\begin{document}
\maketitle

\begin{abstract}
Based on the recently proposed transferable dialogue state generator (TRADE) \cite{wu2019trade} that predicts dialogue states from utterance-concatenated dialogue context, we propose a multi-task learning model with a simple yet effective utterance tagging technique and a bidirectional language model as an auxiliary task for task-oriented dialogue state generation. By enabling the model to learn a better representation of the long dialogue context, our approaches attempt to solve the problem that the performance of the baseline significantly drops when the input dialogue context sequence is long. In our experiments, our proposed model achieves a 7.03\% relative improvement over the baseline, establishing a new state-of-the-art joint goal accuracy of 52.04\% on the MultiWOZ 2.0 dataset.
\end{abstract}

\section{Introduction}
Dialogue state tracking (DST, also known as belief tracking) predicts user's goals in task-oriented dialogue system, where dialogue states are normally represented in the form of a set of slot-value pairs. A variety of approaches to dialogue state tracking are devoted to dealing with two different settings: DST over a predefined domain ontology and DST with slot-value candidates from an open vocabulary. Most of the previous work is based on the first setting, assuming that all possible slot-value candidates are provided in a domain ontology in advance. The task of the dialogue state tracking with this setting is therefore largely simplified to score all predefined slot-value pairs and select the value with the highest score for each slot as the final prediction. Although predefined ontology-based approaches are successfully used on datasets with small ontologies, such as DSTC2 \cite{henderson2014second} and WOZ2.0 \cite{wen2017network}, they are quite limited in both scalability to scenarios with infinite slot values and prediction of unseen slot values.

In order to address these issues of DST over predefined ontologies, recent efforts have been made to predict slot-value pairs in open vocabularies. Among them, TRADE \cite{wu2019trade} proposes to encode the entire dialogue context and to predict the value for each slot using a copy-augmented decoder, achieving state-of-the-art results on the MultiWOZ
2.0 dataset \cite{budzianowski2018multiwoz}. As TRADE simply concatenates all the system and user utterances in previous turns into a single sequence as the dialogue context for slot-value prediction, it is difficult for the model to identify whether an utterance in the dialogue context is from system or user when the concatenated sequence becomes long. We observe that the longest dialogue context after concatenation on the MultiWOZ 2.0 dataset contains 880 tokens. Our experiments also demonstrate that the longer the dialogue context sequence is, the worse TRADE performs.

To deal with this problem, we propose two approaches to modeling long context for better dialogue state tracking. The first method is tagging. While constructing the dialogue context sequence, we insert a tag of \emph{[sys]} symbol in front of each system utterance, and a tag of \emph{[usr]} symbol in front of each user utterance. The purpose of adding such symbolic tags in the concatenated dialogue context sequence is to explicitly enhance the capability of the model in distinguishing system and user utterances. In the second method, we propose to integrate a bi-directional language modeling module into the upstream of the model as an auxiliary task to gain better understanding and representation of the dialogue context. The bi-directional language modeling task is to predict the next word by using forward hidden states and the previous word by using backward hidden states based on the dialogue context sequence without any annotation. With these two approaches, we perform dialogue state tracking in a multi-task learning architecture.


In summary, the contributions of our work are as follows:
\begin{itemize}
    \item We propose a simple tagging method to explicitly separate system from user utterances in the concatenated dialogue context. 
    \item We propose a language modeling task as an auxiliary task to better model long context for DST.
    \item We conduct experiments on the MultiWOZ 2.0 dataset. Both methods achieve significant improvements over the baselines in all evaluation metrics. The joint of the two methods establish a new state-of-the-art results on the MultiWOZ 2.0. In addition, we provide a detailed analysis on the improvements achieved by our methods.
\end{itemize}

\section{Related Work}
Predefined ontology-based DST assumes that all slot-value pairs are provided in an ontology. \citet{mrkvsic2017neural} propose a neural belief tracker (NBT) to leverage semantic information from word embeddings by using distributional representation learning for DST. An extension to the NBT is then proposed by \citet{mrkvsic2018fully}, which learns to update belief states automatically. \citet{zhong2018global} use slot-specific local modules to learn slot features and propose a global-locally self-attentive dialogue state tracker (GLAD). \citet{Nouri2018Toward} propose GCE model based on GLAD by using only one recurrent networks with global conditioning. \citet{ramadan2018large} introduce an approach that fully utilizes semantic similarity between dialogue utterances and the ontology terms. \citet{ren2018towards} propose StateNet which generates a fixed-length representation of the dialogue context and compares the distances between this representation and the value vectors in the candidate set for making prediction. These predefined ontology-based DST approaches suffer from their weak scalability to large ontologies and cannot deal with previously unseen slot values. 

In open vocabulary-based DST, \citet{xu2018end} propose a model that learns to predict unknown values by using the index-based pointer network for different slots. \citet{wu2019trade} apply an encoder-decoder architecture to generate dialogue states with the copy mechanism. However, their method simply concatenates the whole dialogue context as input and does not perform well when the dialogue context is long. We study this problem and propose methods to help the DST model better model long context. Inspired by \citet{zhou-etal-2019-multi} who use an additional language model in question generation, we attempt to incorporate language modeling into dialogue state tracking as an auxiliary task.

\section{Our Methods}
In this section, we describe our proposed methods. First, section \ref{TRADE} briefly introduces the recent TRADE model \cite{wu2019trade} as background knowledge, followed by our methods: utterance tagging in section \ref{data level method} and multi-task learning with language modeling in section \ref{model level method}.

\subsection{Transferable Dialogue State Generator}
\label{TRADE}
\begin{figure*}[htp] 
\centering 
\includegraphics[scale=0.52]{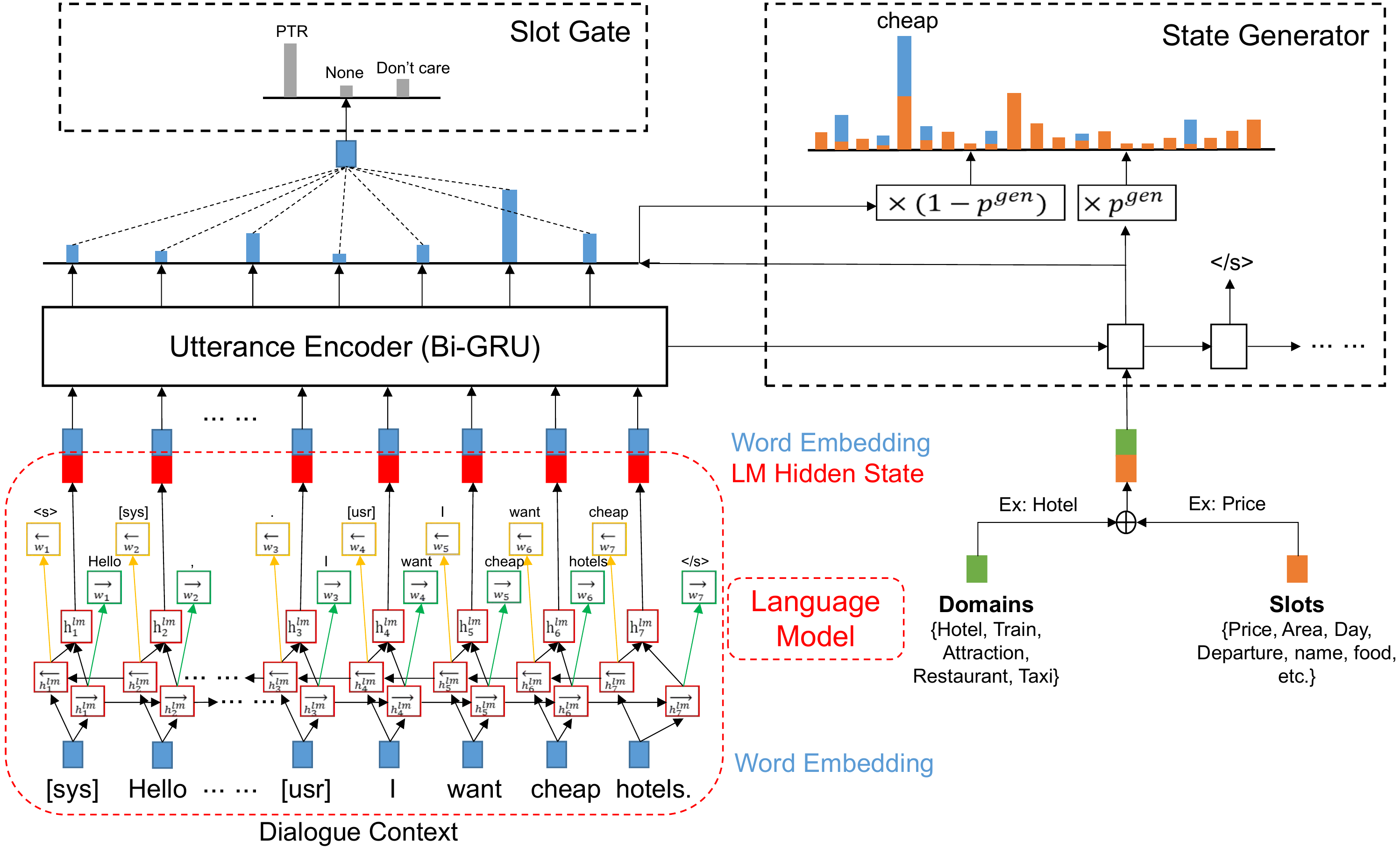}
\caption{Multi-task Learning Framework with Language Modeling Task for Dialogue State Tracking}
\label{LMDST figure} 
\end{figure*}

TRADE is an encoder-decoder model that encodes concatenated previous system and user utterances as dialogue context and generates slot value word by word for each slot exploring the copy mechanism \cite{wu2019trade}. The architecture of TRADE is shown in Figure \ref{LMDST figure} without the language model module. In the encoder of TRADE, system and user utterances in previous dialogue turns are simply concatenated without any labeling. In our experiments, we find that the performance of the TRADE model significantly drops when the length of the dialogue context is long. On the MultiWOZ 2.0 dataset, the maximum length of a dialogue context is up to 880 tokens. About 27\% of instances on the test set have dialogue context sequences longer than 200 tokens. The joint accuracy of the TRADE on these cases drops to lower than 22\%. This suggests that TRADE suffers from long context.

\subsection{Utterance Tagging}
\label{data level method}
To deal with this problem, we first propose a simple method to label system and user utterances by inserting a tag of \emph{[sys]} just at the beginning of each system utterance and a tag of \emph{[usr]} in front of each user utterance when they are concatenated into the dialogue context. We conjecture that mixing system and user utterances in one single sequence may confuse the encoder. It may also mislead the decoder to attend to inappropriate parts and the copy network to copy from wrong utterances. The explicit indicators from the two tags are to help TRADE differ system from user utterances.

\subsection{Multi-task Learning with Language Modeling}
\label{model level method}
We further propose to incorporate a bi-directional language modeling module into the dialogue state tracking model in a multi-task learning framework for DST, which is shown in Figure \ref{LMDST figure}.

The bi-directional language modeling module is to predict the next word and the previous word in the concatenated sequence with the forward and the backward GRU network respectively. We first feed the concatenated dialogue context into the embedding layer. We initialize each word embedding in the dialogue context by concatenating Glove embedding \cite{pennington2014glove} and character embedding \cite{hashimoto2017joint}. This word embedding sequence is then fed into a bi-directional GRU network to get the hidden representations \emph{$\overrightarrow{h_t^{lm}}$} and \emph{$\overleftarrow{h_t^{lm}}$} in two directions, which are used to predict the next and the previous word through a softmax layer as follows:
\begin{align}
  P^{lm}(w_{t+1}|w_{<t+1}) = softmax(W_{f}\overrightarrow{h_t^{lm}}) \\
  P^{lm}(w_{t-1}|w_{>t-1}) = softmax(W_{b}\overleftarrow{h_t^{lm}})
\end{align}

The loss function is defined as the sum of the negative log-likelihood of the next and previous words in the sequence. The language modeling loss \emph{$L^{lm}$} is therefore calculated as follows ($T$ is the length of the concatenated dialogue context sequence):
\begin{equation}
\begin{split}
L^{lm} = &-\sum_{t=1}^{T-1}log(P^{lm}(w_{t+1}|w_{<t+1})) \\
            &- \sum_{t=2}^{T}log(P^{lm}(w_{t-1}|w_{>t-1}))
\end{split}
\end{equation}

The sum of the forward and backward hidden states in the language model module is used as the hidden representation $h_t^{lm}$ for word $w_t$ in the dialogue context: $h_t^{lm} = \overrightarrow{h_t^{lm}} + \overleftarrow{h_t^{lm}}$. We further sum it with the word embedding of $w_t$ and feed the sum into the utterance encoder. Following \citet{wu2019trade}, we include the slot gate and state generator modules in our model and calculate the dialogue state tracking loss \emph{$L^{dst}$}.

The training objective for the multi-task learning framework is to minimize the total loss \emph{$L^{total}$} which is the sum of DST and language modeling loss:
\begin{equation}
L^{total} = L^{dst} + \alpha L^{lm}
\end{equation}
where $\alpha$ is a hyper-parameter which is used to balance the two tasks.

\section{Experiments}
In this section, we evaluated our proposed methods on the public dataset.

\subsection{Datasets \& Settings}
\label{datasets}
We conducted experiments on the MultiWOZ 2.0 \cite{budzianowski2018multiwoz} which is the largest multi-domain task-oriented dialogue dataset, consisting of over 10,000 dialogues from seven domains. Each dialogue is composed of 13.68 turns on average. Following \citet{wu2019trade}, we used five domains excluding \emph{hospital} and \emph{police} domains which account for a small portion and do not appear on the test set.

In our multi-task learning model, both the sizes of hidden states and word embeddings were set to 400. We set the batch size to 8 and applied the delay update mechanism with different step sizes to train the model.


\begin{table}[]
\centering
\resizebox{\textwidth}{!}{%
\scalebox{0.008}{
\begin{tabular}{lcc}
\hline
\textbf{Model} & \textbf{Joint Accuracy} & \textbf{Slot Accuracy} \\ \hline
\multicolumn{3}{c}{\textbf{Baselines}} \\ \hline
GLAD \cite{zhong2018global} & 35.57 & 95.44 \\
TRADE \cite{wu2019trade} & 48.62 & 96.92 \\
COMER \cite{ren2019scalable} & 48.79 & - \\
NADST \cite{Le2020Non-Autoregressive} & 50.52 & - \\
SOM-DST \cite{kim2019efficient} & 51.38 & - \\
DSTQA \cite{zhou2019multi} & 51.44 & 97.24 \\ \hline
\multicolumn{3}{c}{\textbf{Ours}} \\ \hline
Ours & \textbf{52.04} & \textbf{97.26} \\
-LM & 50.15 & 97.10 \\
-Tagging & 51.36 & 97.23 \\ \hline
\end{tabular}
}
}
\caption{Experimental results on the MultiWOZ 2.0 dataset.}
\label{results table}
\end{table}

\begin{figure}[t] 
\centering 
\includegraphics[scale=0.5]{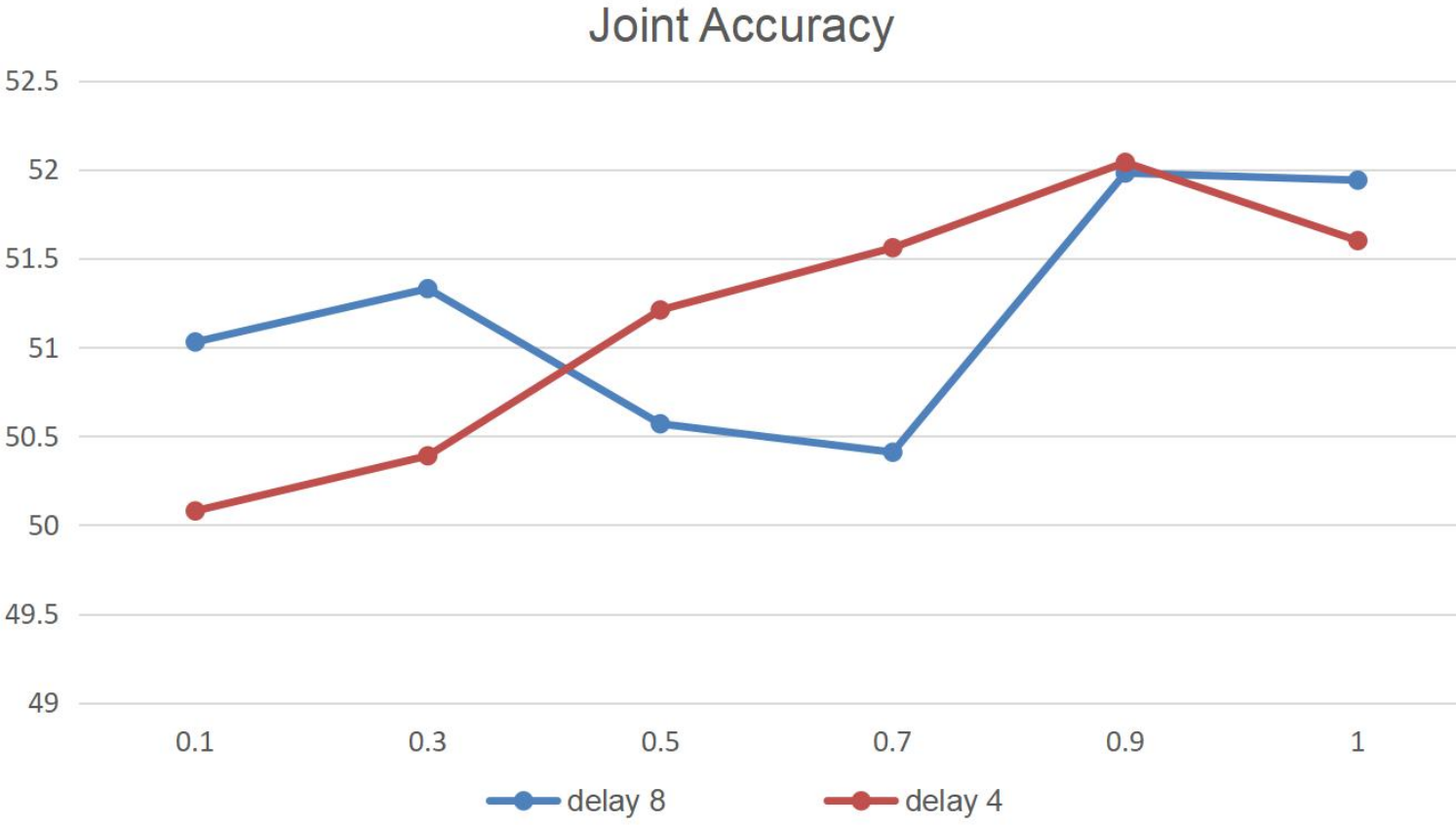}
\caption{The impact of hyper-parameter $\alpha$ and delay update step on DST joint accuracy.}
\label{delay update figure} 
\end{figure}

\subsection{Results}
\label{results and settings}
Joint accuracy and slot accuracy are the two metrics we used to evaluate
the performance on dialogue state tracking. Table \ref{results table} shows the results of our methods and other baselines on the test set of the MultiWOZ 2.0 dataset. Our full model (tagging + language modeling) significantly outperforms several previous state-of-the-art models, including TRADE, and achieves new state-of-the-art results, 52.04\% of joint accuracy and 97.26\% of slot accuracy on the MultiWOZ 2.0. The tagging alone (-LM) can improve the joint accuracy on the MultiWOZ 2.0 by 1.53\% while the auxiliary language modeling (-Tagging) by 2.74\%.

Figure \ref{delay update figure} shows the impact of $\alpha$ and the number of delay update steps on DST. Consequently, our model performs best when we set $\alpha$ to 0.9 and the number of delay update steps to 4.


\subsection{Analysis}
\label{analysis}

\begin{table}[]
\centering
\resizebox{\textwidth}{!}{%
\scalebox{0.0077}{
\begin{tabular}{llllll}
\hline
\multicolumn{1}{c}{\multirow{2}{*}{\textbf{Length}}} & \multicolumn{1}{c}{\multirow{2}{*}{\textbf{Total}}} & \multicolumn{2}{c}{\textbf{Correct Turns}} & \multicolumn{2}{c}{\textbf{Joint Accuracy(\%)}} \\
\multicolumn{1}{c}{} & \multicolumn{1}{c}{} & \multicolumn{1}{c}{\textbf{TRADE}} & \multicolumn{1}{c}{\textbf{Ours}} & \multicolumn{1}{c}{\textbf{TRADE}} & \multicolumn{1}{c}{\textbf{Ours}} \\ \hline
0 - 99 & 2,940 & 2,115 & 2,190 (\textcolor {red}{+75}) & 71.94 & 74.49 (\textcolor{red}{+2.55}) \\
100-199 & 2,466 & 1,028 & 1,129 (\textcolor{red}{+101}) & 41.69 & 45.78 (\textcolor{red}{+4.09}) \\
200-299 & 1,494 & 356 & 445 (\textcolor{red}{+89}) & 23.83 & 29.79 (\textcolor{red}{+5.96}) \\
$\geqslant$ 300 & 468 & 57 & 70 (\textcolor{red}{+13}) & 12.18 & 14.96 (\textcolor{red}{+2.78}) \\ \hline
\end{tabular}

}
}
\caption{Results and statistics on different lengths of dialogue context on the test set.}
\label{lengths statistics table}
\end{table}

\begin{table}[]
\centering
\resizebox{\textwidth}{!}{%
\scalebox{0.0071}{
\begin{tabular}{llllll}
\hline
\multicolumn{1}{c}{\multirow{2}{*}{\textbf{Model}}} & \multicolumn{1}{c}{\multirow{2}{*}{\textbf{Total}}} & \multicolumn{1}{c}{\multirow{2}{*}{\textbf{Correct}}} & \multicolumn{3}{c}{\textbf{Not exactly correct}} \\
\multicolumn{1}{c}{} & \multicolumn{1}{c}{} & \multicolumn{1}{c}{} & \multicolumn{1}{c}{\textbf{Over pred.}} & \multicolumn{1}{c}{\textbf{Partial pred.}} & \multicolumn{1}{c}{\textbf{False pred.}} \\ \hline
TRADE & 7,368 & 3,556 & 791 & 1,480 & 1,541 \\
Ours & 7,368 & \textbf{3,834 (\textcolor{red}{+278})} & 877 (\textcolor{blue}{+86}) & \textbf{1,201 (\textcolor{red}{-279})} & \textbf{1,456 (\textcolor{red}{-85})} \\ \hline
\end{tabular}%
}
}
\caption{Statistics and analysis on different types of prediction errors. The red indicates positive effects, while the blue indicates negative effect.}
\label{errors statistics table}
\end{table}

We further provide a deep analysis on our results on the MultiWOZ 2.0 according to the length of concatenated dialogue context, which are shown in Table \ref{lengths statistics table}. We can clearly observe that the performance of the baseline model drops sharply with the increase of the dialogue context length. We can also find that our model performs better than the baseline in all cases, suggesting that the proposed methods are able to improve modeling long dialogue context for DST.

Table \ref{errors statistics table} shows the statistics of different kinds of prediction errors on the test set of the MultiWOZ 2.0. We define three types of dialogue state prediction errors. Over prediction is that the predicted states not only fully cover the golden states, but also include some redundant slot values. Partial prediction is an error that the predicted states are just part of the golden states with some slot values missing. False prediction denotes that false slot values are predicted for some slots. As shown in Table \ref{errors statistics table}, our model significantly reduces the number of partial and false prediction errors, with the help of better representation of dialogue context.


\section{Conclusion}
In this paper, we have presented the utterance tagging and auxiliary bi-directional language modeling in a multi-task learning framework to model long dialogue context for open vocabulary-based DST. Experiments on the MultiWOZ 2.0 dataset show that our model significantly outperforms the baselines and achieves new state-of-the-art results.

\section*{Acknowledgments}
The present research was supported by the National Natural Science Foundation of China (Grant No. 61861130364) and the Royal Society (London) (NAF$\backslash$R1$\backslash$180122). We would like to thank the anonymous reviewers for their insightful comments. 


\bibliography{acl2020}

\begin{thebibliography}{18}
\expandafter\ifx\csname natexlab\endcsname\relax\def\natexlab#1{#1}\fi

\bibitem[{Budzianowski et~al.(2018)Budzianowski, Wen, Tseng, Casanueva, Ultes,
  Ramadan, and Ga{\v{s}}i{\'c}}]{budzianowski2018multiwoz}
Pawe{\l} Budzianowski, Tsung-Hsien Wen, Bo-Hsiang Tseng, I{\~n}igo Casanueva,
  Stefan Ultes, Osman Ramadan, and Milica Ga{\v{s}}i{\'c}. 2018.
\newblock {M}ulti{WOZ} - a large-scale multi-domain wizard-of-{O}z dataset for
  task-oriented dialogue modelling.
\newblock In \emph{Proceedings of the 2018 Conference on Empirical Methods in
  Natural Language Processing}, pages 5016--5026, Brussels, Belgium.
  Association for Computational Linguistics.

\bibitem[{Hashimoto et~al.(2017)Hashimoto, Tsuruoka, Socher
  et~al.}]{hashimoto2017joint}
Kazuma Hashimoto, Yoshimasa Tsuruoka, Richard Socher, et~al. 2017.
\newblock A joint many-task model: Growing a neural network for multiple nlp
  tasks.
\newblock In \emph{Proceedings of the 2017 Conference on Empirical Methods in
  Natural Language Processing}, pages 1923--1933.

\bibitem[{Henderson et~al.(2014)Henderson, Thomson, and
  Williams}]{henderson2014second}
Matthew Henderson, Blaise Thomson, and Jason~D Williams. 2014.
\newblock The second dialog state tracking challenge.
\newblock In \emph{Proceedings of the 15th Annual Meeting of the Special
  Interest Group on Discourse and Dialogue (SIGDIAL)}, pages 263--272.

\bibitem[{Kim et~al.(2019)Kim, Yang, Kim, and Lee}]{kim2019efficient}
Sungdong Kim, Sohee Yang, Gyuwan Kim, and Sang-Woo Lee. 2019.
\newblock Efficient dialogue state tracking by selectively overwriting memory.
\newblock \emph{arXiv preprint arXiv:1911.03906}.

\bibitem[{Le et~al.(2020)Le, Socher, and Hoi}]{Le2020Non-Autoregressive}
Hung Le, Richard Socher, and Steven~C.H. Hoi. 2020.
\newblock Non-autoregressive dialog state tracking.
\newblock In \emph{International Conference on Learning Representations}.

\bibitem[{Mrk{\v{s}}i{\'c} et~al.(2017)Mrk{\v{s}}i{\'c}, S{\'e}aghdha, Wen,
  Thomson, and Young}]{mrkvsic2017neural}
Nikola Mrk{\v{s}}i{\'c}, Diarmuid~{\'O} S{\'e}aghdha, Tsung-Hsien Wen, Blaise
  Thomson, and Steve Young. 2017.
\newblock Neural belief tracker: Data-driven dialogue state tracking.
\newblock In \emph{Proceedings of the 55th Annual Meeting of the Association
  for Computational Linguistics (Volume 1: Long Papers)}, pages 1777--1788.

\bibitem[{Mrk{\v{s}}i{\'c} and Vuli{\'c}(2018)}]{mrkvsic2018fully}
Nikola Mrk{\v{s}}i{\'c} and Ivan Vuli{\'c}. 2018.
\newblock Fully statistical neural belief tracking.
\newblock In \emph{Proceedings of the 56th Annual Meeting of the Association
  for Computational Linguistics (Volume 2: Short Papers)}, pages 108--113.

\bibitem[{Nouri and Hosseini-Asl(2018)}]{Nouri2018Toward}
Elnaz Nouri and Ehsan Hosseini-Asl. 2018.
\newblock Toward scalable neural dialogue state tracking model.
\newblock In \emph{Advances in neural information processing systems (NeurIPS),
  2nd Conversational AI workshop}.

\bibitem[{Pennington et~al.(2014)Pennington, Socher, and
  Manning}]{pennington2014glove}
Jeffrey Pennington, Richard Socher, and Christopher Manning. 2014.
\newblock Glove: Global vectors for word representation.
\newblock In \emph{Proceedings of the 2014 conference on empirical methods in
  natural language processing (EMNLP)}, pages 1532--1543.

\bibitem[{Ramadan et~al.(2018)Ramadan, Budzianowski, and
  Gasic}]{ramadan2018large}
Osman Ramadan, Pawe{\l} Budzianowski, and Milica Gasic. 2018.
\newblock Large-scale multi-domain belief tracking with knowledge sharing.
\newblock In \emph{Proceedings of the 56th Annual Meeting of the Association
  for Computational Linguistics (Volume 2: Short Papers)}, pages 432--437.

\bibitem[{Ren et~al.(2019)Ren, Ni, and McAuley}]{ren2019scalable}
Liliang Ren, Jianmo Ni, and Julian McAuley. 2019.
\newblock Scalable and accurate dialogue state tracking via hierarchical
  sequence generation.
\newblock In \emph{Proceedings of the 2019 Conference on Empirical Methods in
  Natural Language Processing and the 9th International Joint Conference on
  Natural Language Processing (EMNLP-IJCNLP)}, pages 1876--1885.

\bibitem[{Ren et~al.(2018)Ren, Xie, Chen, and Yu}]{ren2018towards}
Liliang Ren, Kaige Xie, Lu~Chen, and Kai Yu. 2018.
\newblock Towards universal dialogue state tracking.
\newblock In \emph{Proceedings of the 2018 Conference on Empirical Methods in
  Natural Language Processing}, pages 2780--2786.

\bibitem[{Wen et~al.(2017)Wen, Vandyke, Mrk{\v{s}}i{\'c}, Ga{\v{s}}i{\'c},
  Rojas-Barahona, Su, Ultes, and Young}]{wen2017network}
Tsung-Hsien Wen, David Vandyke, Nikola Mrk{\v{s}}i{\'c}, Milica
  Ga{\v{s}}i{\'c}, Lina~M. Rojas-Barahona, Pei-Hao Su, Stefan Ultes, and Steve
  Young. 2017.
\newblock A network-based end-to-end trainable task-oriented dialogue system.
\newblock In \emph{Proceedings of the 15th Conference of the {E}uropean Chapter
  of the Association for Computational Linguistics: Volume 1, Long Papers},
  pages 438--449, Valencia, Spain. Association for Computational Linguistics.

\bibitem[{Wu et~al.(2019)Wu, Madotto, Hosseini-Asl, Xiong, Socher, and
  Fung}]{wu2019trade}
Chien-Sheng Wu, Andrea Madotto, Ehsan Hosseini-Asl, Caiming Xiong, Richard
  Socher, and Pascale Fung. 2019.
\newblock Transferable multi-domain state generator for task-oriented dialogue
  systems.
\newblock In \emph{Proceedings of the 57th Annual Meeting of the Association
  for Computational Linguistics}, pages 808--819, Florence, Italy. Association
  for Computational Linguistics.

\bibitem[{Xu and Hu(2018)}]{xu2018end}
Puyang Xu and Qi~Hu. 2018.
\newblock An end-to-end approach for handling unknown slot values in dialogue
  state tracking.
\newblock In \emph{Proceedings of the 56th Annual Meeting of the Association
  for Computational Linguistics (Volume 1: Long Papers)}, pages 1448--1457.

\bibitem[{Zhong et~al.(2018)Zhong, Xiong, and Socher}]{zhong2018global}
Victor Zhong, Caiming Xiong, and Richard Socher. 2018.
\newblock Global-locally self-attentive encoder for dialogue state tracking.
\newblock In \emph{Proceedings of the 56th Annual Meeting of the Association
  for Computational Linguistics (Volume 1: Long Papers)}, pages 1458--1467.

\bibitem[{Zhou and Small(2019)}]{zhou2019multi}
Li~Zhou and Kevin Small. 2019.
\newblock Multi-domain dialogue state tracking as dynamic knowledge graph
  enhanced question answering.
\newblock \emph{arXiv preprint arXiv:1911.06192}.

\bibitem[{Zhou et~al.(2019)Zhou, Zhang, and Wu}]{zhou-etal-2019-multi}
Wenjie Zhou, Minghua Zhang, and Yunfang Wu. 2019.
\newblock Multi-task learning with language modeling for question generation.
\newblock In \emph{Proceedings of the 2019 Conference on Empirical Methods in
  Natural Language Processing and the 9th International Joint Conference on
  Natural Language Processing (EMNLP-IJCNLP)}, pages 3385--3390, Hong Kong,
  China. Association for Computational Linguistics.

\end{thebibliography}
\bibliographystyle{acl_natbib}




\end{document}